\documentclass[10pt,twocolumn,letterpaper]{article}

\usepackage[pagenumbers]{cvpr} % To force page numbers, e.g. for an arXiv version

\usepackage{graphicx}
\usepackage{amsmath}
\usepackage{amssymb}
\usepackage{pifont}
\usepackage{booktabs}
\usepackage{comment}
\usepackage{commath}
\usepackage[pagebackref,breaklinks,colorlinks]{hyperref}
\usepackage{color,soul}
\usepackage{multirow}

\newcommand{\xmark}{\ding{55}}%
\newcommand{\cmark}{\ding{51}}%
\usepackage[capitalize]{cleveref}
\crefname{section}{Sec.}{Secs.}
\Crefname{section}{Section}{Sections}
\Crefname{table}{Table}{Tables}
\crefname{table}{Tab.}{Tabs.}

\usepackage{amsmath, amsthm, amssymb}

\newtheorem{lemma}{Lemma}[section]

\def\cvprPaperID{3984} % *** Enter the CVPR Paper ID here
\def\confName{CVPR}
\def\confYear{2022}

\begin{document}

\title{ITSA: An Information-Theoretic Approach to Automatic Shortcut Avoidance and Domain Generalization in Stereo Matching Networks}

\author{WeiQin Chuah\footnotemark[1] \hspace{1.5ex} Ruwan Tennakoon\footnotemark[1] \hspace{1.5ex} Reza Hoseinnezhad\footnotemark[1] \hspace{1.5ex} Alireza~Bab-Hadiashar\footnotemark[1] \hspace{1.5ex} David Suter\footnotemark[2]\\
RMIT University, Australia\footnotemark[1] \hspace{3.5ex} Edith Cowan University~(ECU), Australia\footnotemark[2]\\
{\tt\small \{wei.qin.chuah,ruwan.tennakoon,rezah,abh\}@rmit.edu.au, d.suter@ecu.edu.au}}

\maketitle

%%%%%%%%% ABSTRACT
\begin{abstract}
   State-of-the-art stereo matching networks trained only on synthetic data often fail to generalize to more challenging real data domains. In this paper, we attempt to unfold an important factor that hinders the networks from generalizing across domains: through the lens of shortcut learning. We demonstrate that the learning of feature representations in stereo matching networks is heavily influenced by synthetic data artefacts (shortcut attributes). To mitigate this issue, we propose an Information-Theoretic Shortcut Avoidance~(ITSA) approach to automatically restrict shortcut-related information from being encoded into the feature representations. As a result, our proposed method learns robust and shortcut-invariant features by minimizing the sensitivity of latent features to input variations. To avoid the prohibitive computational cost of direct input sensitivity optimization, we propose an effective yet feasible algorithm to achieve robustness. We show that using this method, state-of-the-art stereo matching networks that are trained purely on synthetic data can effectively generalize to challenging and previously unseen real data scenarios. Importantly, the proposed method enhances the robustness of the synthetic trained networks to the point that they outperform their fine-tuned counterparts (on real data) for challenging out-of-domain stereo datasets. 
\end{abstract}

%%%%%%%%% BODY TEXT
\section{Introduction}
\label{sec:intro}
Stereo matching is a fundamental task in computer vision and is widely used for depth sensing in various applications such as augmented reality~(AR), robotics and autonomous driving. In recent years, end-to-end trained Convolutional Neural Networks~(CNNs) have achieved impressive results for this task as quantified by the performance on several publicly available stereo-matching benchmarks~\cite{chang2018pyramid, guo2019group, kendall2017end, xu2020aanet, zhang2019ga}. 

\begin{figure}
    \centering
    \subfloat[KITTI-15]{\includegraphics[width=0.15\textwidth,height=0.28\textwidth]{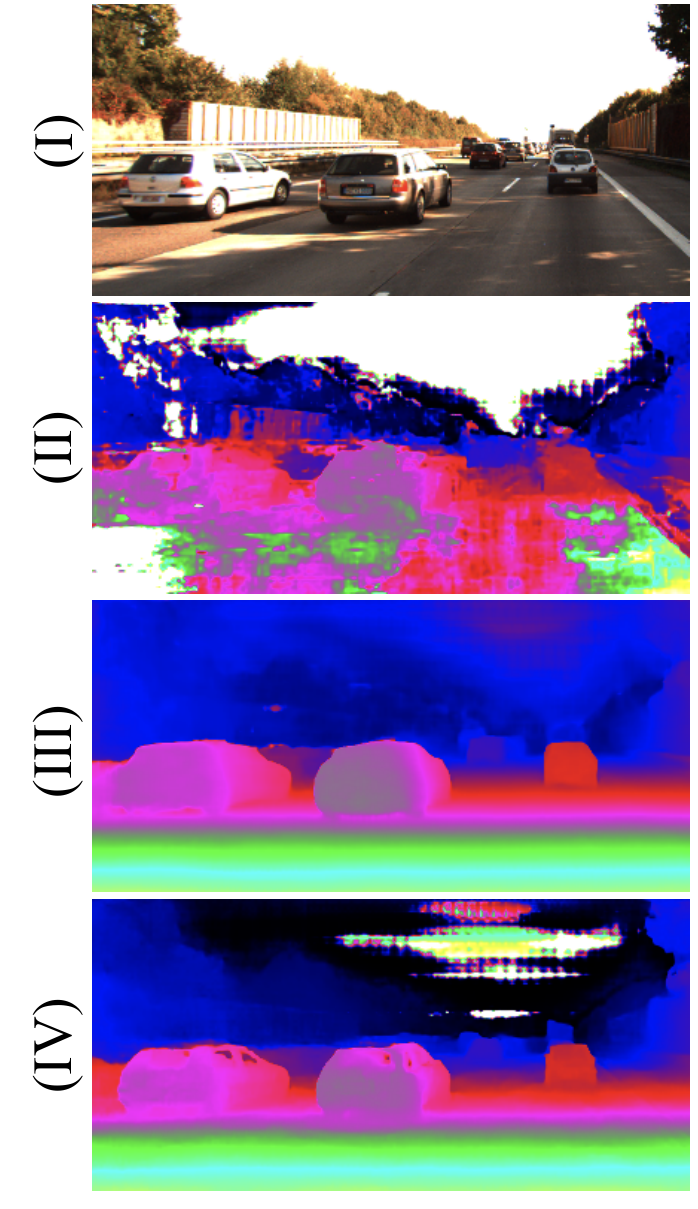}}
    \subfloat[DrivingStereo]{\includegraphics[width=0.13\textwidth,height=0.28\textwidth]{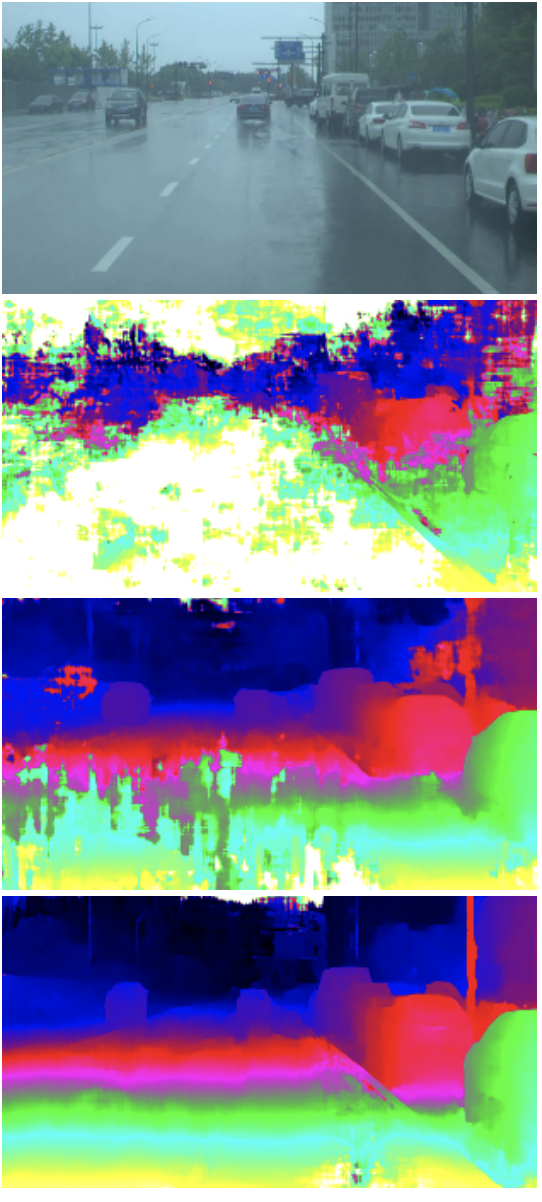}}
    \subfloat[Oxford]{\includegraphics[width=0.11\textwidth,height=0.28\textwidth]{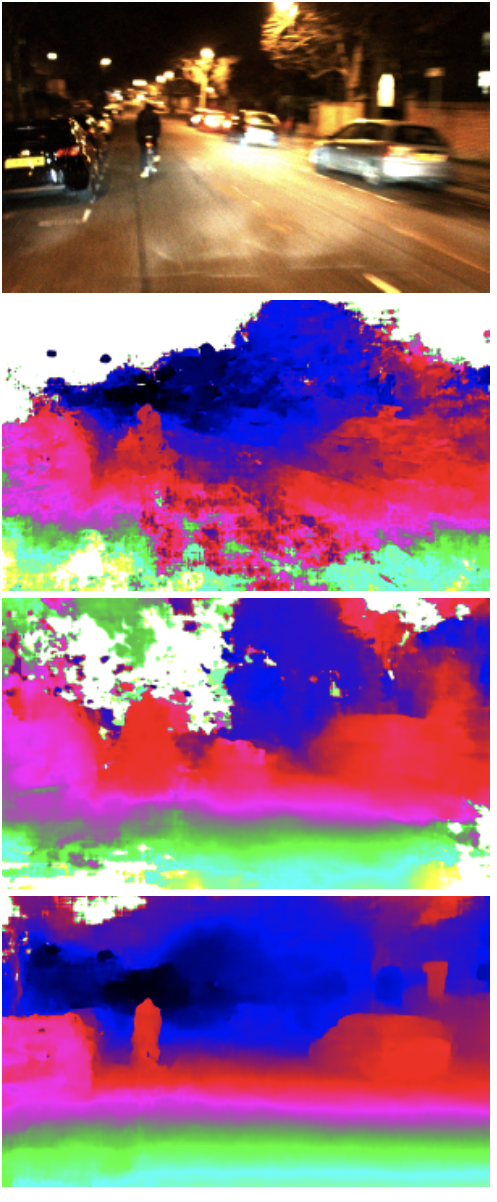}}
    \subfloat[Middlebury]{\includegraphics[width=0.1\textwidth,height=0.28\textwidth]{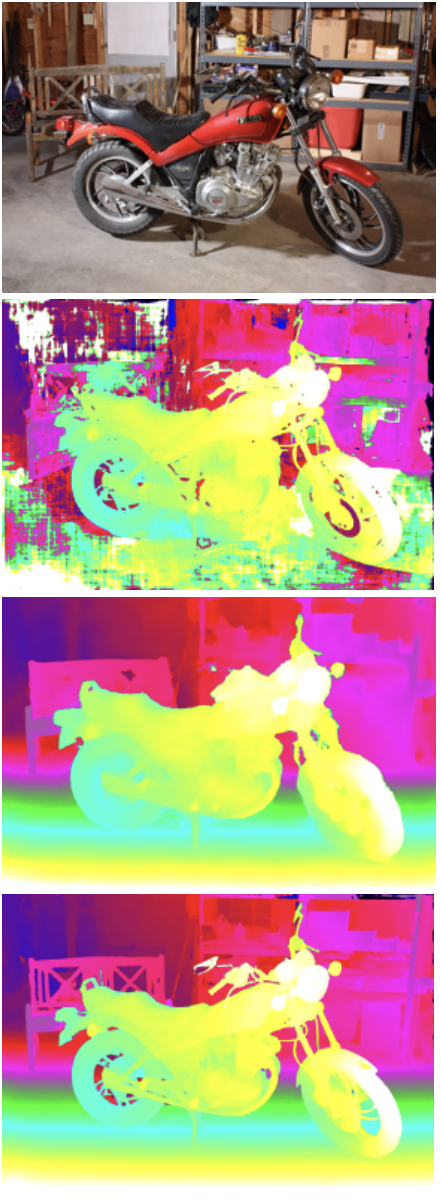}}
    \caption{Comparison of disparity maps estimated by PSMNet~\cite{chang2018pyramid} when it is trained under different settings and across multiple domains. Each column shows the results for a realistic domain namely: KITTI~2015~\cite{Menze2018JPRS}, DrivingStereo~\cite{yang2019drivingstereo}, Oxford Robotcar~\cite{RobotCarDatasetIJRR} and Middlebury~\cite{scharstein2014high}. Rows from top to bottom show a sample image (I) , the prediction for the Scene Flow pre-trained model (II), KITTI-15 fine-tuned model (III), and the proposed ITSA optimized method (IV). Comparing these figures shows that PSMNet trained solely on synthetic data performs poorly on real data and fine tuning only improves the result for KITTI dataset (still fails to generalize for other scenarios). The proposed method performs well across the board (best viewed in color).}
    \label{fig:motivation}
\end{figure}

Generally, end-to-end stereo-matching networks require a large amount of labelled data for training. To overcome this challenge, many state-of-the-art networks are initially trained on labelled synthetic data, commonly generated using game engines. However, models trained using synthetic data do not generalize well to unseen realistic domains. For example, the PSMNet~\cite{chang2018pyramid} pre-trained on the Scene Flow dataset~\cite{mayer2016large} performs poorly when tested on unseen realistic domains as illustrated in~\cref{fig:motivation}. Therefore, in practice, the networks trained with synthetic data are fine-tuned using labelled data from the relevant target domain. However, collecting even a relatively small amount of dense ground truth data in the real-world can be challenging for tasks like stereo-matching~\cite{tonioni2019learning, liu2020stereogan}. Furthermore, to be practically useful in many applications, a stereo-matching model should be able to generalize effortlessly to different domains like day and night times, varying weather conditions, etc. Collecting data for fine-tuning that cover all possible situations is both difficult and expensive. It is therefore highly desirable to remove the fine-tuning requirement. 

It is known that neural networks, including stereo matching networks, can learn superficial shortcut features (or spurious correlations with the target labels), which prevent them from generalizing across different domains~\cite{geirhos2020shortcut,Beery_2018_ECCV}. We found that stereo matching networks trained on synthetic data are susceptible to exploiting shortcuts in synthetic data such as (1)~consistent local statistics~(RGB color features) between the left and right stereo images and (2)~over-reliance on local chromaticity features~(e.g. color, illumination, texture) of the reference stereo viewpoint. Detailed analysis and discussion are included in~\cref{sec:shortcuts}. Dependency on these shortcut cues, instead of the desirable semantic and structural representations, means that these networks would fail drastically when the spurious correlations between shortcuts and labels do not exist in a new (unseen) domain~\cite{recht2019imagenet}. While several shortcut-removal approaches have been previously proposed~\cite{hendrycks2019robustness, carlucci2019domain, shi2020informative}, most of these methods are manually designed ~(e.g. carefully selected data augmentations~\cite{hendrycks2019robustness, carlucci2019domain}) and rely on the assumption that the shortcuts could be identified in advance. However, shortcuts can be non-intuitive, task-specific, and difficult to identify~\cite{dagaev2021too, minderer2020automatic}. 

Our goal is to train a stereo matching network on synthetic data that can generalize to realistic scenes without the need for fine-tuning. To achieve this, we propose an information-theoretic approach to automatically restrict the shortcut-related information from being encoded from the input into the feature representations. The approach is based on the well known information bottleneck (IB) principle that proposes to optimize the following objective \cite{tishby2015deep,alemi2017deepvib}: 
\begin{equation}
    \underset{\theta}{\textup{argmax}~}{I\left( Y, Z; \theta \right) - \beta   I\left( X, Z; \theta \right)}
    \label{eqn:IB}
\end{equation} 
where $Z$ is the encoding of input $X$, $Y$ is the target, $I$ is mutual information and $\beta \in [0,~1]$ is the hyperparameter that controls the size of the information bottleneck. While optimizing the IB objective leads to compressed feature representations, our empirical experiments showed that these compressed features are neither robust nor shortcut-invariant~(details are provided in~\cref{sec:robust-ib}). Consequently, the IB optimized networks may still incorporate shortcuts and remain fragile when tested in unseen domains. 
The recently introduced robust IB criterion~\cite{pensia2020extracting} encourages the learning of both robust and compressive features by replacing the mutual information in IB with statistical Fisher information. %Fisher information measures the sensitivity of the distribution of the extracted features with respect to changes that occur in the inputs.
%Therefore, in our approach, we use this robust IB criterion to learn a generalisable stereo matching model.  the application of IB with mutual information as a measure
Robust IB is presented in the context of learning features that are robust to adversarial attacks and to the best of our knowledge it has not been used for domain generalization. 

In our approach, we combine the task loss~(e.g. smooth L1 loss) with Fisher information to learn a generalizable stereo matching model. 
Although such an objective can work in theory, straightforward optimization of the Fisher information by gradient descent requires computation of the second-order derivatives and is therefore computationally expensive for tasks with high dimensional inputs such as stereo matching and semantic segmentation. %To overcome this shortcoming, we propose a loss term that approximates the Fisher information loss. 
To overcome this shortcoming, we propose ITSA which consists of a novel loss term and perturbation technique to approximate the optimization of the Fisher information loss.
The proposed ITSA is computationally efficient, and as we show by extensive experiments, it can promote the learning of shortcut-invariant features. Unlike the existing domain-invariant stereo matching networks~\cite{zhang2020domain, Shen_2021_CVPR}, the proposed ITSA does not involve significant network alteration and is model-agnostic. Therefore, as shown in the experiments section, it can be easily integrated with different stereo matching networks. 

The empirical results show that stereo-matching networks trained on synthetic data, with the proposed ITSA, can generalise to realistic data without fine-tuning. 
Additional experiments on challenging out-of-domain stereo datasets~(e.g. different adverse weathers and night scenes) show that our method also improves the overall robustness of the stereo matching networks and importantly even outperforms the networks fine-tuned on realistic domains when tested on these challenging datasets.
The main contributions of this paper include:
\begin{itemize}
    \item We show that learning feature representations that are less sensitive to input variations can significantly enhance the synthetic to realistic domain generalization, and robustness in stereo matching networks.
    \item We introduce a novel loss function that enables us to minimize the Fisher information, without computing the second-order derivatives.
    \item We also show that the application of the proposed framework is not limited to stereo matching task, and can be used in training models for non-geometry based vision problems such as semantic segmentation.
\end{itemize}
The rest of the paper is organized as follows. \cref{sec:related-works}~describes the related work in the field of learning-based stereo matching networks, domain generalization and shortcut learning. \cref{sec:methods}~presents the proposed method for automatic shortcut avoidance and domain generalization. Experimental results and discussions are presented in~\cref{sec:experiments}, and \cref{sec:conclusion} concludes the paper.

%-------------------------------------------------------------------------

\section{Related Work} \label{sec:related-works}

\noindent \textbf{Learning-based Stereo Matching Networks} \\
In recent years, end-to-end learned deep stereo matching networks have excelled in most datasets and benchmarks~\cite{chang2018pyramid, kendall2017end, xu2020aanet, zhang2019ga}. These networks generally have three sub-modules (1) feature extraction sub-network, (2) cost-volume generator, and (3) cost aggregation and refinement sub-network. There are two main types of stereo matching networks based on how the cost volume is generated. 

\textit{Correlation-based stereo matching networks} construct the cost volume by correlating the features extracted from the two views. %A series of 2D convolution layers followed by up-sampling modules are employed for cost regularization and disparity estimation. 
Previously proposed correlation-based methods include DispNetC~\cite{mayer2016large}, iResNet~\cite{liang2018learning}, CRL~\cite{pang2017cascade}, SegStereo~\cite{yang2018segstereo}, and AANet~\cite{xu2020aanet}. Although these methods are usually computationally efficient, semantics and structural information in the feature representations are lost due to the correlation operation~\cite{guo2019group}. As a result, the correlation-based stereo matching methods usually have inferior performance compared to the concatenated-based methods. %they lose information due to correlation operation and usually has inferior performance. WHAT DOES IT MEAN???

%\noindent
\textit{Concatenation-based methods} use a cost volume that is a simple assembly of features extracted from the two views. 
Examples of the state-of-the-art concatenation-based stereo matching networks include PSMNet~\cite{chang2018pyramid}, GANet~\cite{zhang2019ga}, GCNet~\cite{kendall2017end}, StereoDrNet~\cite{chabra2019stereodrnet} and EMCUA~\cite{nie2019multi}. While these networks can achieve superior performance in stereo matching, they require labelled samples from the target environments, for fine-tuning. Without fine-tuning, these networks cannot generalize to unseen test data. 

To overcome this problem, Zhang~\etal~\cite{zhang2020domain} proposed DSMNet, which employs Domain Normalization and non-local graph-based filtering layers to enforce the learning of structural features that are domain-invariant. Similarly, Shen~\etal~\cite{Shen_2021_CVPR} introduced CFNet, an efficient network architecture with multi-scale cost volume fusion and refinement, to enforce the learning of robust structural representation for stereo matching. In contrast, we have identified shortcut learning~\cite{geirhos2020} as a major factor that hinders stereo matching networks from generalizing across domains. In this work, we show that avoiding shortcut learning can effectively enhance the robustness of the stereo matching networks and enables a model to generalize across domains. This is evidenced by showing networks' superior performance on challenging realistic data without fine-tuning. \\

\noindent \textbf{Single Domain Generalization} \\
Domain generalization typically involves forcing DNNs to learn domain-invariant features, using data sampled from multiple source domains~\cite{li2018domain, qiao2020learning}. On the other hand, single domain generalization is a more challenging problem because only one source domain is available for training. To solve this problem, Volpi~\etal~\cite{VolpiNSDMS18} proposed adversarial data augmentation~(ADA), which aims to expand and diversify the distribution of training data. Specifically, ADA creates “fictitious” yet “challenging” new populations, simulating data sampled from novel domains, using adversarial training. In a similar fashion, Qiao~\etal~\cite{qiao2020learning} proposed a novel framework that employs ADA and meta-learning to enforce the learning of domain-invariant features. 
While these works focus on minimizing the domain differences, we are interested in learning robust and shortcut-invariant features that are transferable across different domains. To this end, we propose ITSA, an information-theoretic approach to prevent shortcut learning~(see next section), particularly in the stereo matching networks.\\

\noindent \textbf{Shortcut Learning} \\
Geirhos \etal~\cite{geirhos2020shortcut} coined the term shortcut learning as a phenomenon where DNNs learn trivial solutions by relying on superficial features (shortcuts). These features are spuriously correlated with the target labels, without contributing to transferability across contexts. For example, image classification networks tend to rely on shortcuts such as backgrounds~\cite{geirhos2020shortcut, Beery_2018_ECCV} and textures~\cite{geirhos2018imagenet, wang2018learning} to improve their performance. However, these networks fail to generalize to unseen domains, where the spurious correlations between shortcuts and labels are violated~\cite{recht2019imagenet}. Similarly, we obeserved that stereo matching networks trained on synthetic data also have a tendency to exploit shortcuts to produce accurate depth results in synthetic domains. Consequently, these networks fail drastically when tested in unseen realistic environments.

Several attempts have been made to restrict the learning of identified shortcuts and generalize DNNs across domains~\cite{shi2020informative,carlucci2019domain,choi2021robustnet,hendrycks2019robustness,wang2018learning}. These methods reply on having some shortcut-related prior knowledge and usually include data augmentations~\cite{hendrycks2019robustness, carlucci2019domain}, whitening transformation~\cite{choi2021robustnet} or dropout-based regularization~\cite{shi2020informative} as part of their solutions. However, shortcuts are non-trivial, task-specific, and are often difficult to be identified a priori~\cite{dagaev2021too, minderer2020automatic}. In contrast, our proposed method automatically avoid shortcut learning without requiring shortcut-related knowledge in advance.%In contrast, our proposed method automatically avoid shortcut learning and does not require any shortcut-related knowledge in advance. 

%-------------------------------------------------------------------------
\section{Methodology} \label{sec:methods}
% Figure: overview method. 
\begin{figure*}[t]
    \centering
    \includegraphics[width=0.75\textwidth]{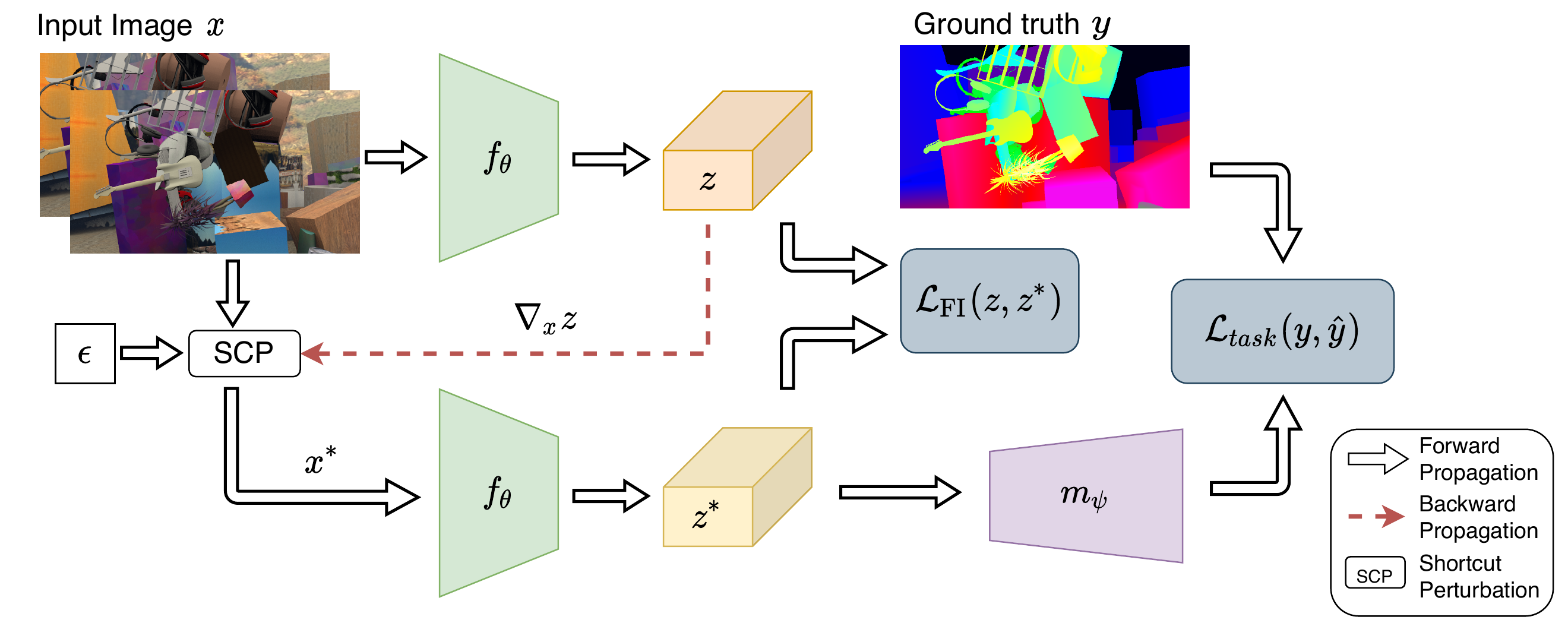}
    \caption{An overview of the proposed shortcut-avoidance strategy to achieve domain generalization in stereo matching networks. The parameters are shared across the two feature extractor networks~$f_\theta$~(best viewed in color).}
    \label{fig:overview}
\end{figure*}

\subsection{Problem Definition}
In this work, we focus on the synthetic-to-realistic domain generalization for stereo matching. Given a synthetic stereo data set $D_{syn}$ consisting of stereo image pairs $\left \{ x^{(i)}_{syn,l}, x^{(i)}_{syn,r} \right \}_{i=1}^n$ with corresponding ground-truth~disparity~$\left \{ y^{(i)}_{syn}\right \}_{i=1}^n$, the goal is to design a robust and shortcut-invariant stereo matching network that can accurately predict disparity map~$\hat{y}^{(i)}$ for unseen realistic environments~$D_{real}$. 

Our approach to achieve synthetic-to-realistic domain generalization is to use an information-theoretic measure to automatically restrict the shortcut-related information from being included in feature representations.  

\subsection{Model}
\label{sec:model}
A typical stereo-matching network can be represented by the following equation:
\begin{small}
\begin{equation}
    \hat{y}^{(i)} = m_\psi \left (\mathbb{C}\left ( f_\theta \left ( x^{(i)}_{l} \right ), f_\theta \left ( x^{(i)}_{r} \right ) \right )  \right )
\end{equation}
\end{small}
where $f_\theta\left( \cdot \right)$ is the feature extraction sub-network, $\mathbb{C}\left ( \cdot \right)$ the cost volume and $m_\psi \left ( \cdot \right)$ the cost aggregation and refinement sub-network. The refined cost volumes are converted to disparity maps~$\hat{y}$ via the soft argmin~\cite{kendall2017end} operation.

Our proposed method~(ITSA) can be applied to any stereo-matching network that has the above structure. In the experiments section, we show the result of applying the proposed algorithm to different stereo-matching networks with concatenation cost (we observed similar results with correlation-based methods) volumes~\cite{chang2018pyramid, guo2019group, Shen_2021_CVPR}. The high-level structure of the network including the proposed shortcut avoidance strategy is shown in \cref{fig:overview}.

\subsection{Loss function}
Our main contribution is the loss function devised to automatically restrict the shortcut-related information from being encoded in the learning process. As we explained earlier, the information bottleneck~(IB) principle~\cite{tishby2015deep,alemi2017deepvib} is typically used to compress features and would be a natural choice to achieve this objective.

The standard $\mathcal{L}_\text{IB}$ loss defined in~\cref{eqn:IB}, which uses mutual information to quantify information content, was designed to extract features that are both concise and relevant for prediction. However, models trained by this loss are not robust to existence of artefacts that can generate shortcuts (similar to adversarial distortions mentioned in~\cite{pensia2020extracting}). 

To demonstrate the above point, we conducted a toy experiment. In this experiment, we investigated the efficacy of using IB loss for helping digit recognition networks~(DRNs) to generalize from MNIST (source)~\cite{lecun1998gradient} to MNIST-M~\cite{ganin2016domain} (target) dataset. The former contains images of handwritten digits with black background, and the latter is created by combining the MNIST digits with randomly extracted from color patches as their background.
All networks were trained on the MNIST training set only and the top-1 accuracy~($\%$) was employed for evaluation. The details of the experiment are included in the supplementary document. As shown in~\cref{tab:toy}, the standard IB can effectively reduce over fitting, and achieves the best performance in the source domain. However, it fails to generalize its performance to the unseen domain. Importantly, it even performs worse than the baseline networks in the unseen target domain. 

\begin{table}[t]
\centering
\resizebox{0.3\textwidth}{!}{%
\footnotesize
\begin{tabular}{lcc}
\hline
Method    & MNIST~\cite{lecun1998gradient} & MNIST-M~\cite{ganin2016domain} \\ \hline
ERM       & 97.9 $\pm$ 0.14                & 40.9  $\pm$ 2.95           \\
IB~\cite{alemi2017deepvib}        & \textbf{99.0} $\pm$ 0.47       & 21.8  $\pm$ 0.21           \\
RIB~\cite{pensia2020extracting}       & 98.3 $\pm$ 0.13                & 52.8  $\pm$ 1.04  \\ 
ITSA      & 98.1 $\pm$ 0.38                & \textbf{56.9}  $\pm$ 1.23           \\ \hline
\end{tabular}
}
\caption{Performance comparison of digit recognition networks optimized via empirical risk minimization~(ERM), the information bottleneck~(IB)~\cite{alemi2017deepvib}, its robust variant~(RIB)~\cite{pensia2020extracting} and our proposed method~(ITSA). While IB performs well in the in-domain tests, it performs poorly on out-of-domain tests.  %\textcolor{red}{According to the CVPR guidelines, the captions appear to be at the bottom for figures and tables.}
}
\label{tab:toy}
\end{table}

\subsubsection{Robust Information Bottleneck and Fisher Information} \label{sec:robust-ib}
As our aim is to develop an IB based cost function that is not susceptible to existence of shortcuts in source data, we take inspiration from the robust IB principle~\cite{pensia2020extracting}. Robust IB utilizes the statistical Fisher information~$\Phi(Z|X)$ of the extracted features~$Z$ parameterized by the inputs~$X$ as a more robust measure of information (in place of $I(Z,X)$). The Fisher information~$\Phi(Z|X)$ is defined as:
\begin{small}
\begin{equation}
    \Phi(Z|X)=\int_\mathcal{X} \Phi(Z|X=x) p_X(x)dx ,  \label{eqn:fish}
\end{equation}
\end{small}
where 
\begin{small}
\begin{equation}
    \Phi(Z|X=x) = \int_\mathcal{Z} \norm{\nabla_x \log{p_{Z|X}(z|x)}}_2^2 p_{Z|X}(z|x)dz.
    \label{eqn:fish_}
\end{equation}
\end{small}
The term $\Phi(Z|X=x)$ in Eq.~(\ref{eqn:fish}, \ref{eqn:fish_}) can be regarded as the sensitivity of the latent distribution $p_{Z|X}(\cdot|x)$, with respect to changes at the input~$x$. Therefore, optimizing the Fisher information,~$\Phi(Z|X)$, will minimize the average sensitivity of the latent distribution with respect to change of inputs~$X$. 
%Consequently, robust features that are also invariant to changes in the inputs can be sampled from the estimated latent distribution. 
As shortcuts are generated by data artefacts that are transient~\footnote{We use transient to describe image attributes that are inconsistent across domains, and spuriously correlated with the true label. These features may include backgrounds, textures, image style, etc.} by nature, they are sensitive to perturbations of input data~\cite{geirhos2020shortcut}. As such, minimizing the Fisher information is a step towards promoting the learning of shortcut-invariant features. Our conjecture is supported by the results of the toy experiment included in~\cref{tab:toy}. The DRNs constrained by the Fisher information~(RIB) achieved better performance than the IB networks in the target domain.

In order to minimize the Fisher information expressed in~\cref{eqn:fish_}, one has to compute second order derivatives such as $\nabla_\theta\nabla_x \log{p_{Z|X}(z|x)}$, which is computationally prohibitive for tasks with large dimensional inputs such as stereo matching, semantic segmentation, etc.~\cite{shi2021gradient}. To overcome this issue, we propose ITSA, a simple yet computationally feasible approach to promote the learning of shortcut-invariant features.

\subsubsection{Approximating Fisher information}
%Optimizing the Fisher information measure defined in~\cref{eqn:fish} is related to minimizing the average sensitivity of the latent distribution~$p_{Z|X}$ with respect to changes in the input $X$ and minimising $\Phi\left ( Z \mid X=x \right )$ on average would lead to minimising $\Phi\left ( Z \mid X \right )$. 
Optimizing the Fisher information~$\Phi\left ( Z \mid X \right )$ measure defined in~\cref{eqn:fish} is related to minimizing~$\Phi\left ( Z \mid X=x \right )$. By adding a regularization term such as $\Phi\left ( Z \mid X=x \right )$ to the loss function, we can penalize the transient features and discourages networks from learning shortcuts. To calculate this term, we employ a first order approximation as described below.  

\begin{lemma}
\label{lem:lem1}
If $\epsilon > 0$, $u$ is a unit vector (i.e. $\left \|  u \right \| = 1$, we refer to as the shortcut perturbation) and $x^*=x+\epsilon u$, then, subject to first order approximation:
\begin{small}
\begin{equation}
\begin{split}
    \Phi\left ( Z \mid X=x \right ) = \frac{\mathbb{E}_{z}\left [ \left | p_{Z \mid X=x^*}\left( z \right) - p_{Z \mid X=x}\left( z \right)\right | \right ]^2}{\epsilon^2 \cos^2{\psi}}  \\+ \mathcal{V}\left [ \left \| \nabla_x \log p_{Z \mid X=x}\left( z \right) \right \|_2 \right ]
    \end{split}
    \label{equ:Fapprox}
\end{equation}
\end{small}
where $\mathbb{E}_z\left[ \upsilon \right]$ and $\mathcal{V}\left[ \upsilon \right]$ are the expectation and variance of $\upsilon$, and $\psi$ is the angle between $u$ and $\nabla_x p_{Z \mid X=x}$. 
\end{lemma}
\noindent Proof is given in the supplementary material.

The first term in the RHS of \cref{equ:Fapprox} will be minimized when the divergence (distance) between the two distributions, $p_{Z \mid X=x}$ and $p_{Z \mid X=x+\epsilon u}$, is reduced. There are many popular divergence measures between distributions, such as Kullback-Leibler divergence, Jensen-Shannon divergence, Total Variation, the Wasserstein distance, etc. In this work, we choose the Wasserstein distance: as the distributions $p_{Z \mid X=x}$ and $p_{Z \mid X=x+\epsilon u}$ may not have common supports and it leads to a simpler loss function.

%In the case of a deterministic feature extractor, the second term in the RHS of \cref{equ:Fapprox} will be zero and $z = f_\theta\left( x \right)$. Thus, the distributions $p_{Z\mid X=x}$ and $p_{Z\mid X=x^*}$ can be seen as two degenerate distributions (i.e. Dirac delta distributions) located at points $z = f_\theta\left( x \right)$ and $z^* = f_\theta\left( x +\epsilon u \right)$. In this case, the Wasserstein-$p$ distance can be simplified as:
In the case of a deterministic feature extractor, which is common in stereo matching networks, the distributions $p_{Z\mid X=x}$ and $p_{Z\mid X=x^*}$ can be seen as two degenerate distributions (i.e. Dirac delta distributions) located at points $z = f_\theta\left( x \right)$ and $z^*=f_\theta\left( x^* \right)$. Furthermore, the $\mathcal{V}\left [ \cdot \right]$ in~\cref{equ:Fapprox} will be zero. In this case, the Wasserstein-$p$ distance can be simplified as:
\begin{small}
\begin{equation}
    W_p(p_{Z\mid X=x^*},p_{Z\mid X=x}) = \left( \left \| z^* - z \right \|^p_2 \right)^{1/p}.
\end{equation}
\end{small}

%Furthermore, the second term in the RHS of \cref{equ:Fapprox} will be zero for a deterministic feature extractor.

Using the above insights, we can see that minimizing $ \left \| z^* - z \right \|_2 $ is a step towards minimizing $\Phi\left ( Z \mid X=x \right )$ (for $p=1$). Thus, we propose to promote the learning of robust and shortcut-invariant features in stereo matching networks, by optimizing the overall loss function defined below:%we define the overall loss function for training the stereo-matching network to be:
\begin{small}
\begin{equation}
    \mathcal{L} = \mathcal{L}_{smooth_{L1}}\left( \hat{y}, y \right) + \frac{\lambda}{2} \left( \mathcal{L}_{\text{FI}}\left(z_l, z^*_l \right) +  \mathcal{L}_{\text{FI}}\left(z_r, z^*_r \right)\right)
    \label{equ:overallLoss}
\end{equation}
\end{small}
where $\hat{y}$ and $y$ are the estimated and ground-truth disparity maps, $\mathcal{L}_{\text{FI}}$ is our proposed Fisher information loss function defined as:
\begin{small}
\begin{equation}
    \mathcal{L}_{\text{FI}} = \sum_{i=1}^n\left \| z^{(i)} - z^{* (i)} \right \|_2
\end{equation}
\end{small}
and $\mathcal{L}_{smooth_{L1}}$ is the smooth-L1 loss function commonly employed for optimizing stereo matching networks~\cite{chang2018pyramid,guo2019group,zhang2019ga,zhang2020domain}.

\subsubsection{Shortcut Perturbation~(SCP)} \label{sec:scp}
In order to compute~$\mathcal{L}_\text{FI}$, we need to define $u$ (refer to as shortcut perturbation and is introduced in~\cref{lem:lem1}): $u = \frac{\nabla_x z^{(i)}}{\left \| \nabla_x z^{(i)} \right \|_2}$ where~$\nabla_x z^{(i)}$ is the gradient of the extracted features~$z$ with respect to input. The shortcut-perturbed image can then be expressed as:
\begin{small}
\begin{equation}
    x^{*(i)} = x^{(i)} + \epsilon \frac{\nabla_x z^{(i)}}{\left \| \nabla_x z^{(i)} \right \|_2}
    \label{equ:ImagePertubation}
\end{equation}
\end{small}
%An overview of the construction of the shortcut-perturbed input images is illustrated in~\cref{fig:overview_perturb}.
The above perturbation will put more weight on pixels that are sensitive to changes in the input. Intuitively, pixels with large absolute value of~$\nabla_x z$ will have significant impact in altering the statistics of encoded latent distributions and the extracted latent feature representations. Moreover, these pixels are also likely to include shortcuts as shortcuts are highly sensitive to perturbations of the input~\cite{geirhos2020shortcut}.

To examine the accuracy of the above approximations, we trained the digit recognition network of our toy experiment with the proposed SCP and $\mathcal{L}_\text{FI}$~(ITSA). As the proposed method is specifically designed for domain generalization, our method can effectively generalize the network to unseen domains and achieve better performance~($4\%$) than the robust information bottleneck as shown in~\cref{tab:toy}.
\begin{figure}
    \centering
    \includegraphics[width=0.45\textwidth]{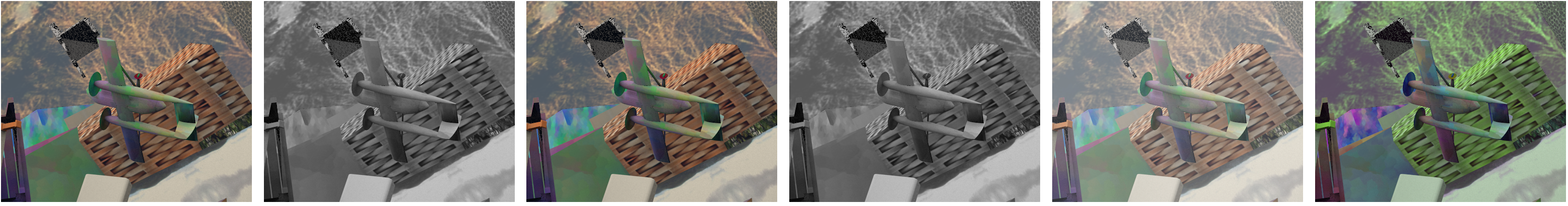}
    \includegraphics[width=0.45\textwidth]{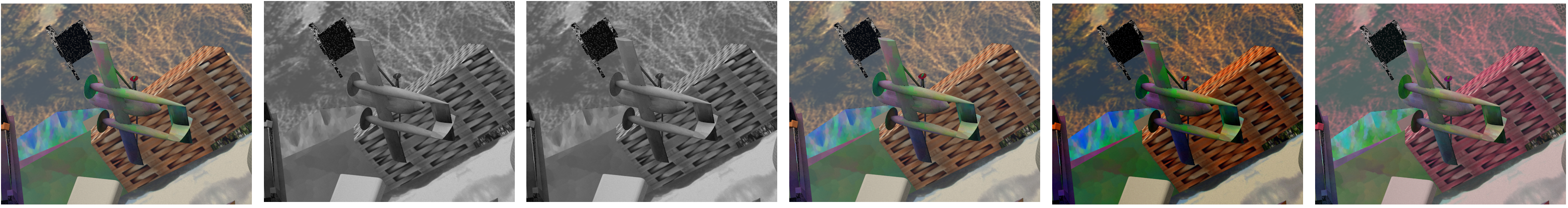}
    \includegraphics[width=0.45\textwidth]{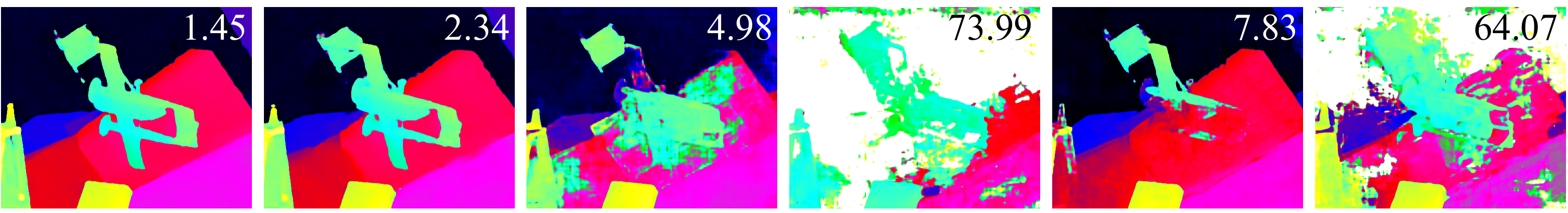}
    \includegraphics[width=0.45\textwidth]{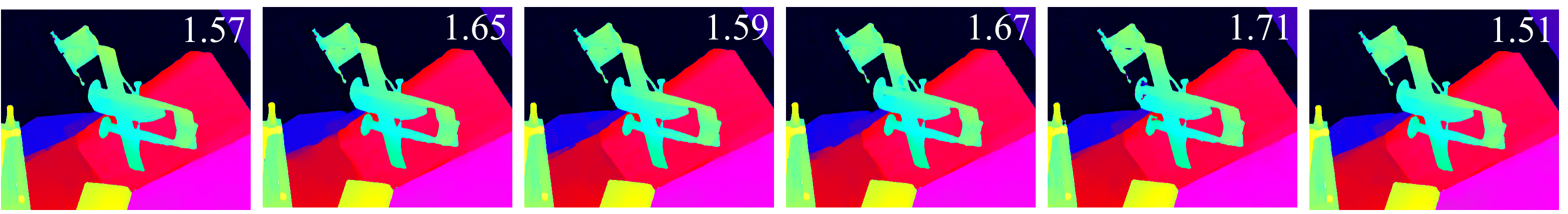}
    \caption{Examples of shortcuts in stereo matching networks. The left and right input images are included in the top two rows. The disparity maps estimated by the baseline PSMNet~\cite{chang2018pyramid} are included in the third row and ITSA-PSMNet in the bottom row. The performance of the baseline PSMNet deteriorates substantially when the shortcut attributes are distorted or removed from the input stereo images. The corresponding EPE is displayed on the estimated disparity map. Best viewed in color and zoom in for details.}
    \label{fig:shortcuts}
\end{figure}

%-------------------------------------------------------------------------
\section{Experiments} \label{sec:experiments}
\subsection{Experimental Settings} \label{sec:implementations}
\noindent \textbf{Datasets and Metrics:}
\textit{Scene Flow}~\cite{mayer2016large} is a large collection of synthetic stereo images with dense disparity ground truth. It contains FlyingThings3D, Driving and Monkaa subsets, and provides 35,454 training and 4,370 testing images. In our experiments, all stereo matching networks are trained on the Scene Flow dataset \emph{only}.  

The realistic datasets used in our experiments include \textit{KITTI2012}~\cite{Geiger2012CVPR} and \textit{KITTI2015}~\cite{Menze2018JPRS} containing 193 and 200 stereo images of outdoor driving scenes,  \textit{Middlebury}~\cite{scharstein2014high} containing 15 images of high resolution indoor scenes, and \textit{ETH3D}~\cite{schoeps2017cvpr} containing 27 low resolution, greyscale stereo images of both indoor and outdoor scenes. Furthermore, datasets covering different weather conditions provided by the \textit{DrivingStereo}~\cite{yang2019drivingstereo} dataset, and night-time provided by \textit{Oxford Robotcar}~\cite{RobotCarDatasetIJRR}) were also included to evaluate the robustness of our proposed method. All the above datasets come with with sparse ground truth. 

% Metric
We evaluated the performance of disparity estimation using the D1 error rate~(\%), with different pixel threshold. The D1 metric computes the percentage of bad pixels~(disparity end-point error larger than the threshold) in the left frame. Following the advice of data originators, a threshold of 3 pixels is selected for KITTI and DrivingStereo, 2 pixels for Middlebury, and 1 pixel for ETH3D. 

\noindent \textbf{Baselines \& Implementation Details:} \label{Sec:Implementation}
We have selected three popular and top-performing stereo matching networks namely PSMNet~\cite{chang2018pyramid}, GwcNet~\cite{guo2019group} and CFNet~\cite{Shen_2021_CVPR} as the baseline networks for our experiments. We have selected these networks mainly due to the fact that PSMNet and GwcNet are well-studied, and commonly employed as a baseline in many prior works~\cite{wang2019pseudo, you2019pseudo, zhang2019adaptive}; and CFNet is one of the recently proposed state-of-the-art stereo matching networks. The networks are implemented using PyTorch framework and are trained end-to-end with Adam $(\beta_1=0.9, \beta_2=0.999)$ optimizer. Similar to the original implementations of the selected networks, our data processing includes color normalization and random cropping the input images to size $H=256$ and $W=512$. Following the original implementation of CFNet, asymmetric chromatic augmentation and asymmetric occlusion~\cite{yang2019hierarchical} are also employed for data augmentation in CFNet. The maximum disparity for PSMNet and GwcNet is set to 192, and for CFNet is set to 256. All models are trained from scratch for 20 epochs with learning rate set to 0.001 for the first 10 epochs and decreased by half for another 10 epochs. The batch size is set to 12 for training on 2 NVIDIA RTX 8000 Quadro GPUs. The models are trained using \textbf{synthetic data only} and directly tested using data from different realistic datasets. For all experiments included in the following sections, the hyper-parameters $\lambda$ and $\epsilon$ were set to $0.1$ and $0.5$ respectively. The hyper-parameter tuning experiments are detailed in the supplementary document. The code of our implementations is available at: \url{https://anonymous.4open.science/r/ITSA-D281} %\textit{\{URL hidden for anonymity\}}.

\begin{table}[t]
\centering
\resizebox{0.4\textwidth}{!}{%
\footnotesize
\begin{tabular}{c|ccc}
\hline
Inputs & PSMNet~\cite{chang2018pyramid} & GwcNet~\cite{guo2019group} & CFNet~\cite{Shen_2021_CVPR} \\ \hline
No Augment ($x$)         & 1.38 & 0.85 & 1.00 \\ \hline
ACJ        & 13.98 & 3.13 & 1.34 \\
%ARP         & 2.02 & 1.24 & 1.02 \\
GrayScale (L) & 37.68 & 8.41 & 1.32 \\
GrayScale (R) & 9.82 & 2.25 & 1.09 \\
SCP         & 5.84 & 2.90 & 2.55 \\ \hline
%$x^* = x + \epsilon \nabla_x z$   & 3.52 & 1.70 & 1.72 \\ \hline
\end{tabular}
}
\caption{Analysis of the effect of data augmentation on the performance of stereo matching networks. All networks are only trained on the Scene Flow training set and the EPE metric is employed for evaluation. The results show that removing shortcut related artefacts (by data augmentation) negatively impact the performance of these networks. In particular, our proposed augmentation can even significantly impact robust methods (e.g. CFNet).}
\label{tab:efficacy}
\end{table}

\subsection{Shortcuts in stereo matching networks} \label{sec:shortcuts}
Our hypothesis is that the baseline stereo matching networks naively trained on synthetic data only, learn to exploit common artefacts of synthetic stereo images as shortcut features. These artefacts include (1)~consistent local statistics~(RGB color features) between between the left and right stereo images and (2)~over-reliance  on  local  chromaticity features of the reference stereo viewpoint. 

To empirically verify the above, we tested three baseline networks trained \textit{only} with synthetic data (i.e. Scene flow), using augmented stereo inputs images. The augmented stereo images were derived from the Scene Flow test set using the following strategies: (1)~Chromatic Augmentation~(e.g.~asymmetrical color jittering~(ACJ)~\cite{yang2019hierarchical} and gray scaling) %and (2)~Asymmetric Random Patching (ARP) - perturbs several local patches positioned randomly in the left or right image by changing color or additive Gaussian noise, 
and (2) the shortcut-perturbation~(SCP, explained in~\cref{sec:scp}). If a network has learnt to utilize the transient attributes (related to shortcut), distorting those in the input space will negatively impact its performance. Experimental results, given in~\cref{tab:efficacy}, showed that using these augmented images as inputs has substantially worsened the performance of the stereo matching networks. 

Interestingly, the SCP images also deteriorate the performance of the best performing robust stereo matching networks such as CFNet~\cite{Shen_2021_CVPR}. In~\cref{sec:results}~and~\ref{sec:robust}, we show that our method can enhance the robustness of CFNet and significantly improve its performance in unseen realistic environments and anomalous scenarios. 

The qualitative results, shown in~\cref{fig:shortcuts}, demonstrate that the performance of baseline networks (third row) deteriorated significantly when the color features consistency between stereo viewpoints is violated. Moreover, as shown in the fourth column of~\cref{fig:shortcuts}, removing the chromaticity features from the reference image will causes substantial performance reduction in the baseline networks. In contrast, our proposed method reduces the exploitation of shortcut features and shows better robustness to adverse data augmentation scenarios, without using these shortcut-related knowledge~(see last row of~\cref{fig:shortcuts}). 

\begin{table}[t]
\centering
\resizebox{0.4\textwidth}{!}{%
\footnotesize
\begin{tabular}{cc|cc|cc}
\hline
\multirow{2}{*}{{SCP}} & \multirow{2}{*}{{$\mathcal{L}_\text{FI}$}} & {PSMNet}               & {GwcNet}               & {PSMNet}               & {GwcNet}               \\ 
                              &                                & \multicolumn{2}{c|}{{KITTI-2012}}                      & \multicolumn{2}{c}{{KITTI-2015}}                      \\ \hline
\xmark                             & \xmark                              & 27.4                          & 11.7                          & 29.3                          & 12.8                          \\
\cmark                             & \xmark                              & 8.1                             & 5.3                             & 8.6                             & 5.9                             \\
\cmark                              & \cmark                               & \textbf{5.2} & \textbf{4.9} & \textbf{5.8} & \textbf{5.4} \\ \hline
\end{tabular}}
\caption{Ablation results on PSMNet~\cite{chang2018pyramid} and GwcNet~\cite{guo2019group}. SCP is the proposed shortcut perturbations and $\mathcal{L}_\text{FI}$ is the proposed loss function in \cref{equ:overallLoss}. The D1 metric was used for evaluation.} 
\label{tab:lambda}
\end{table}

\subsection{Ablation Study} \label{sec:ablation}
This section presents the results of our study on the efficacy of each component of the proposed method. We first trained the baseline networks with the proposed shortcut-perturbation augmentation~(SCP) \textit{only}. %The inputs were perturbed according to \cref{equ:ImagePertubation}. 
Next, we trained the baseline networks with both the shortcut-perturbed stereo-images and the proposed loss function in equation~\cref{equ:overallLoss}~(SCP + $\mathcal{L}_\text{FI}$). All the networks in this section were only trained with the synthetic Scene Flow dataset. %The hyperparameter $\lambda$ and $\epsilon$  were set to $0.1$ in all these experiments. 

As shown in~\cref{tab:lambda}, the baseline networks, trained only with synthetic data, perform poorly when tested on KITTI data. The performance improved when shortcut-perturbations~(SCP) were used in the training stage for input image augmentations. Further improvement in both baseline networks can be seen when using the proposed method i.e. SCP with the proposed loss function. We have not included CFNet in the ablation study as it is specifically designed for synthetic to real domain generalization.

\begin{table}[t]
\centering
\resizebox{0.45\textwidth}{!}{%
\footnotesize
\begin{tabular}{l|cc|ccc|c}
\hline
\multicolumn{1}{c|}{\multirow{2}{*}{Methods}} & \multicolumn{2}{c|}{KITTI}  & \multicolumn{3}{c|}{Middlebury}              & \multirow{2}{*}{ETH3D} \\
\multicolumn{1}{c|}{}                         & 2012         & 2015         & Full          & Half          & Quarter      &                        \\ \hline \hline
HD$^3$~\cite{yin2019hierarchical}                         & 23.6         & 26.5         & 50.3          & 37.9          & 20.3         & 54.2                   \\
PSMNet~\cite{chang2018pyramid}                                       & 27.4         & 29.3         & 60.4          & 29.1          & 19.6          & 16.1                   \\
GwcNet~\cite{guo2019group}                                       & 11.7         & 12.8         & 45.5          & 18.1          & 10.9         & 9.0                    \\
CasStereo~\cite{gu2020cascade}                                     & 11.8         & 11.9         & 40.6          & -             & -            & 7.8                    \\
GANet~\cite{zhang2019ga}                                         & 10.1         & 11.7         & 32.2          & 20.3          & 11.2         & 14.1                   \\ \hline
DSMNet~\cite{zhang2020domain}                                        & 6.2          & 6.5          & 21.8 & 13.8          & \textbf{8.1} & 6.2                    \\
CFNet~\cite{Shen_2021_CVPR}                                         & 4.7          & 5.8          & 28.2          & 13.5          & 9.4         & 5.8                    \\ \hline
ITSA-PSMNet                                   & 5.2          & 5.8          & 28.4          & 12.7          & 9.6         & 9.8                    \\
ITSA-GwcNet                                   & 4.9          & 5.4          & 26.8          & 11.4          & 9.3         & 7.1                    \\
ITSA-CFNet                                    & \textbf{4.2} & \textbf{4.7} & \textbf{20.7}          & \textbf{10.4}          & 8.5          & \textbf{5.1}           \\ \hline
\end{tabular}} 
\caption{Synthetic-to-realistic domain generalization evaluation using KITTI, Middlebury and ETH3D training sets. All methods are trained on the Scene Flow dataset and directly tested on the three real datasets. Pixel error rate with different threshold are employed: KITTI 3-pixel, Middlebury 2-pixel and ETH3D 1-pixel.}
\label{tab:cross-domain}
\end{table}

\subsection{Synthetic-to-Realistic Domain Generalization Evaluation} 
\begin{figure}[t]
    \centering
    \subfloat[Input]{\includegraphics[width=0.115\textwidth]{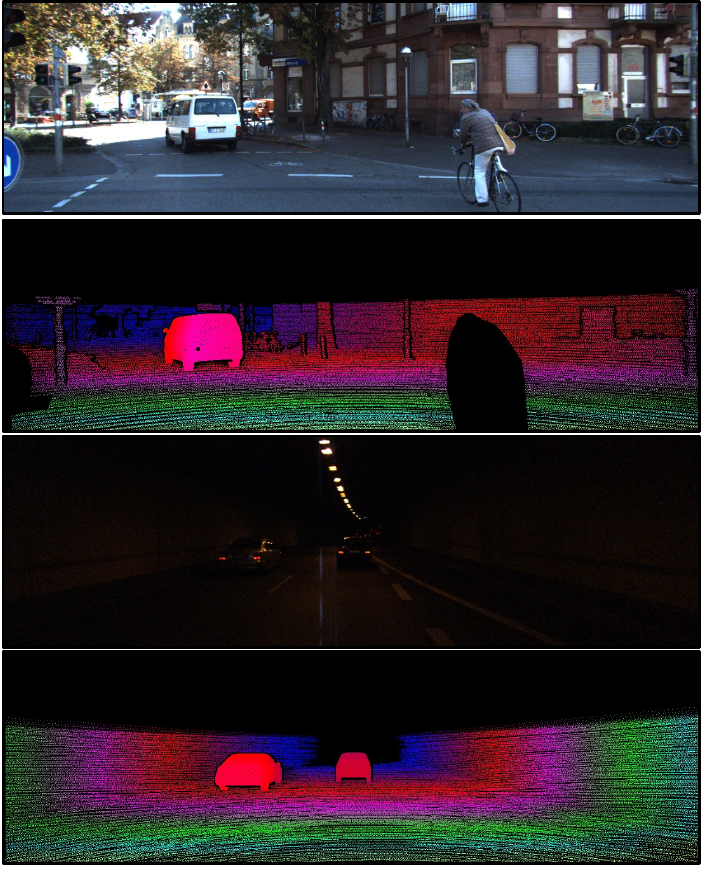}}
    %\hspace{0.05mm}
    \subfloat[PSMNet~\cite{chang2018pyramid}]{\includegraphics[width=0.115\textwidth]{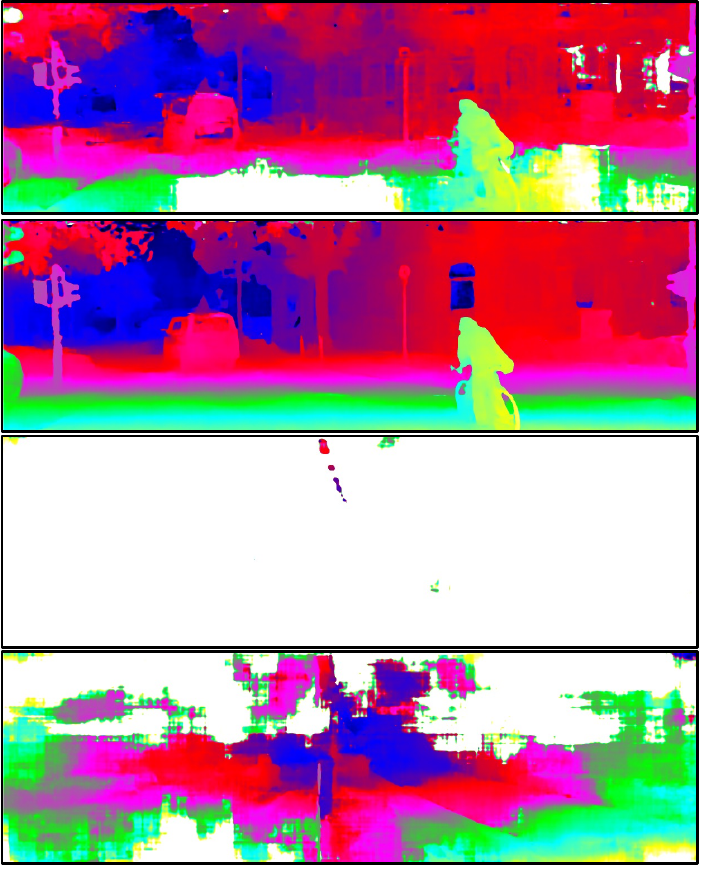}}
    %\hspace{0.05mm}
    \subfloat[GwcNet~\cite{guo2019group}]{\includegraphics[width=0.115\textwidth]{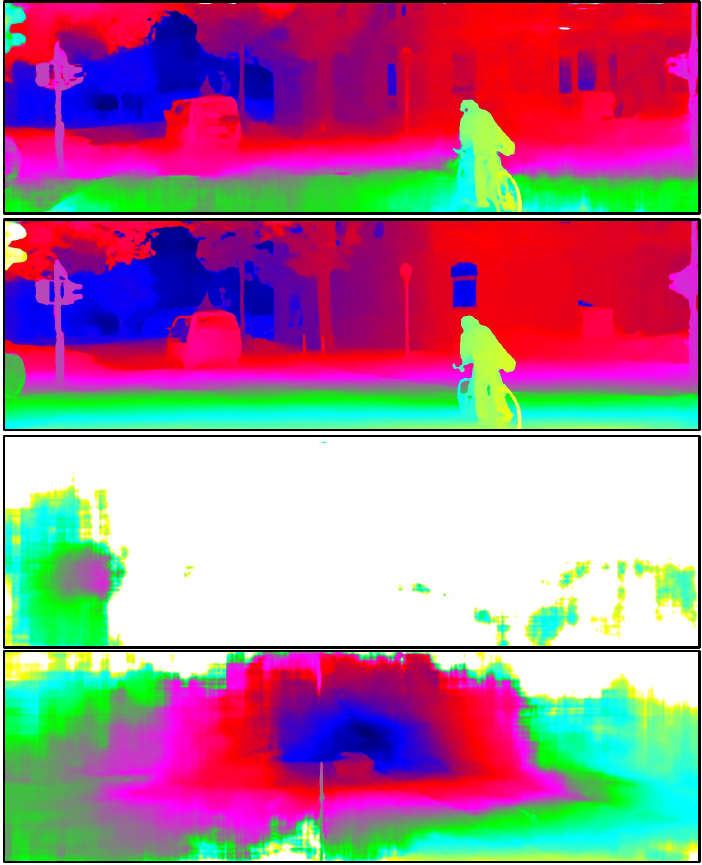}}
    %\hspace{0.05mm}
    \subfloat[CFNet~\cite{Shen_2021_CVPR}]{\includegraphics[width=0.115\textwidth]{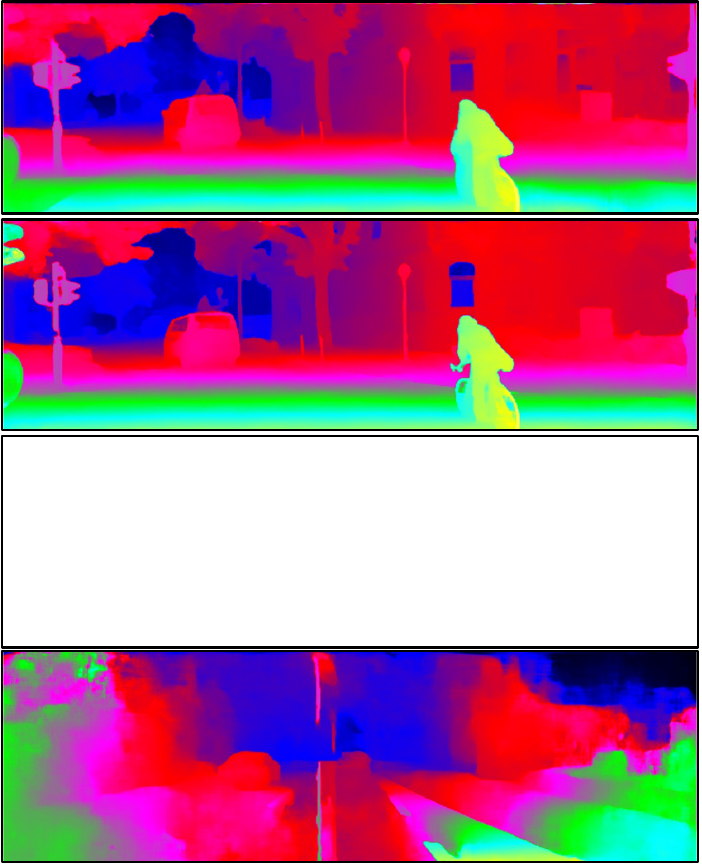}}
    \caption{Qualitative results on KITTI 2015 stereo data. For each example, the results of the baseline networks are presented on the top row and the results from our method are included in the bottom row. The corresponding left image and ground truth are included in column~(a). Our method can significantly improve the stereo matching performance even in scenario with poor lighting condition. Best viewed in color and zoom in for details.}
    \label{fig:cross_domain}
\end{figure}

In~\cref{tab:cross-domain}, we compare the synthetic-to-realistic domain generalization performance of our method with the state-of-the-art stereo matching networks~\cite{yin2019hierarchical, chang2018pyramid, guo2019group, gu2020cascade,zhang2019ga, zhang2020domain,Shen_2021_CVPR} on the four realistic datasets. \textit{All networks are trained on the synthetic Scene Flow training set only}. We found that the proposed ITSA substantially improved the domain generalization performance~($6.8\%-23.5\%$) of the selected stereo networks~(PSMNet~\cite{chang2018pyramid} and GwcNet~\cite{guo2019group}), outperforming the state-of-the-art stereo matching networks in the realistic datasets. The improved networks also outperform DSMNet~\cite{zhang2020domain} on the KITTI 2012~\cite{Geiger2012CVPR} and KITTI~2015~\cite{Menze2018JPRS} datasets, and achieve comparable performance as the CFNet on the Middlebury~\cite{scharstein2014high} datasets. 
In addition, we show that ITSA is even capable of further enhancing the robustness and cross-domain performance of CFNet~\cite{Shen_2021_CVPR}, which was the best performing stereo matching networks in the Robust Vision Challenge 2020. Comparison of qualitative results generated by the baseline networks and our methods are included in~\cref{fig:cross_domain}.

\subsection{Robustness to Anomalous Scenarios} \label{sec:robust}
Here, we analyze the robustness to anomalous conditions of a network trained on synthetic data with the proposed ITSA. The anomalous conditions include night-time, foggy and rainy weather conditions. In this comparison, we train the same network twice: (1) pre-train using synthetic data followed by fine-tuning on realistic KITTI~2015 dataset (common strategy), (2) train only using synthetic data with the proposed SCP and $\mathcal{L}_\text{FI}$~(ITSA). We also included the pre-trained counterpart of CFNet~\cite{Shen_2021_CVPR} to illustrate the efficacy of our method in further enhancing the network robustness. 

In~\cref{tab:robust}, we show that the fine-tuned~(FT) networks generally has better performance when tested on data similar to the KITTI training data~(sunny and cloudy). In contrast, our method~(ITSA) can substantially improve the robustness and overall performance of the PSMNet~\cite{chang2018pyramid} and GwcNet~\cite{guo2019group}, without using the real-world data.
The overall performance performance of fine-tuned CFNet is slightly better than its ITSA counterpart. However, as mentioned earlier, the proposed ITSA improves CFNet performance when only using synthetic data for training. 
The results demonstrate that our method could effectively improve the robustness and performance of existing stereo matching networks, and extends these networks to real-world applications, without using the real data for fine-tuning. 

\label{sec:results}

\begin{table}[t]
\centering
\resizebox{0.45\textwidth}{!}{%
\footnotesize
\begin{tabular}{c|c|c|ccccc|c}
\hline
Models & FT & ITSA & Sun & Cloud & Rain & Fog & Night & Avg \\ \hline \hline
\multirow{2}{*}{PSMNet~\cite{chang2018pyramid}} & \cmark & \xmark  & \textbf{3.94} & \textbf{2.82} & 11.51 & 6.50 & 16.66 & 8.28 \\
 & \xmark & \cmark  & 4.78 & 3.24 & \textbf{9.43} & \textbf{6.31} & \textbf{8.56} & \textbf{6.46} \\ \hline
\multirow{2}{*}{GwcNet~\cite{guo2019group}} & \cmark & \xmark & \textbf{3.10} & \textbf{2.46} & 12.34 & 5.98 & 25.33 & 9.84 \\
 & \xmark & \cmark & 4.35 & 3.31 & \textbf{9.78} & \textbf{5.88} & \textbf{9.41} & \textbf{6.55} \\ \hline
\multirow{3}{*}{CFNet~\cite{Shen_2021_CVPR}} & \xmark & \xmark & 4.89 & 4.64 & 10.74 & 5.43 & 16.19 & 8.38 \\
 & \cmark & \xmark & \textbf{1.79} & \textbf{1.65} & \textbf{5.20} & \textbf{1.59} & 11.56 & \textbf{4.36} \\
 & \xmark & \cmark & 3.42 & 2.87 & 5.32 & 4.32 & \textbf{8.95} & 4.98 \\ \hline
\end{tabular}}
\caption{Robustness evaluation on anomalous scenarios. Our method~(ITSA) consistently enhances the robustness of selected stereo matching networks and outperform the fine-tuned~(FT) models in the real-world anomalous scenarios including rainy and foggy weather and night-time. The performances were evaluated using the D1 metric.}
\label{tab:robust}
\end{table}

\subsection{Extension to Semantic Segmentation}
% Include qualitative and quantitative results on semantic segmentation. 
Similar to stereo matching networks, semantic segmentation networks trained on synthetic data also fail to generalize to realistic data~\cite{qiao2020learning, yue2019domain}. Here, we show that the proposed ITSA can easily be extended to the semantic segmentation task to promote the learning of shortcut-invariant feature and enhance domain generalization. We have selected the commonly employed FCN~\cite{long2015fully} paired with ResNet-50
as the baseline network. The network was trained on synthetic GTAV~\cite{richter2016playing} dataset only and evaluated on real Cityscapes~\cite{Cordts2016Cityscapes} dataset. The mean intersection over union~(mIoU) metric was employed for performance evaluation. As shown in~\cref{tab:segmentation}, the proposed method~(ITSA) can also improve the synthetic-to-realistic domain generalization performance of semantic segmentation networks and achieve comparable performance with the existing domain generalization methods~(IBN-Net~\cite{pan2018two}, ISW~\cite{choi2021robustnet} and DRPC~\cite{yue2019domain}). This further demonstrates the effectiveness of our proposed method in promoting shortcut-invariant features and enhancing the performance of domain generalization. The implementation details and qualitative results of our method are included in the supplementary document.

\begin{table}[t]
\centering
\resizebox{0.45\textwidth}{!}{%
\footnotesize
\begin{tabular}{cc|cc|cc|cc}
\hline
\multicolumn{2}{c|}{IBN-Net~\cite{pan2018two}} & \multicolumn{2}{c|}{ISW~\cite{choi2021robustnet}} & \multicolumn{2}{c|}{DRPC~\cite{yue2019domain}} & \multicolumn{2}{c}{ITSA} \\ \hline
22.17 & \multirow{2}{*}{7.47 $\uparrow$} & 28.95 & \multirow{2}{*}{\textbf{7.63} $\uparrow$} & 32.45 & \multirow{2}{*}{4.97 $\uparrow$} & 28.71 & \multirow{2}{*}{6.65 $\uparrow$} \\
\multicolumn{1}{l}{29.17} &  & \multicolumn{1}{l}{36.58} &  & \multicolumn{1}{l}{\textbf{37.42}} &  & \multicolumn{1}{l}{35.36} &  \\ \hline
\end{tabular}}
\caption{Synthetic-to-realistic domain generalization performance comparison on semantic segmentation task. All networks were trained on the GTAV~\cite{richter2016playing} synthetic dataset only and evaluated on the Cityscapes~\cite{Cordts2016Cityscapes} validation set~(G $\to$ C). The mean intersection over union~(mIoU) metric was employed for evaluation.}
\label{tab:segmentation}
\end{table}

\section{Limitations}
Although the proposed method can significantly improve the performance of stereo matching networks without fine-tuning and even outperform their fine-tuned counterpart when tested in unseen challenging environments (e.g. rain and night-time), its performance remains fragile under extreme conditions (e.g. heavy rain and extreme low light), which may occur in real scenarios. This is reflected by the large errors reported in~\cref{tab:robust}. By looking at samples with large errors, we noticed that those inaccuracies are largely due to having insufficient light source, lens glare/flares and reflection on specular surfaces (wet grounds). In our future work, we aim to address these issues and develop a robust system that can handle these extreme scenarios in real-world applications. 

%-------------------------------------------------------------------------
\section{Conclusion} \label{sec:conclusion}
In this work, we have presented ITSA: a novel information theory-based approach for domain generalization in stereo matching networks. To address the shortcut learning challenge, we propose to minimize the sensitivity of the extracted feature representations to the input perturbations, measured via the Fisher information. We further proposed an efficient algorithm to optimize the Fisher information objective. Experimental results show that the proposed method consistently promotes the learning of robust and shortcut-invariant features, and substantially enhances the performance of existing stereo matching networks in cross-domain generalization, even outperforming their fine-tuned counterparts in challenging scenarios. We also show that the proposed method can be easily extended for non-geometry based vision problems such as semantic segmentation.

%%%%%%%%% REFERENCES
{\small
\bibliographystyle{ieee_fullname}
\bibliography{cvpr_draft}

\begin{thebibliography}{10}\itemsep=-1pt

\bibitem{alemi2017deepvib}
Alex Alemi, Ian Fischer, Josh Dillon, and Kevin Murphy.
\newblock Deep variational information bottleneck.
\newblock In {\em ICLR}, 2017.

\bibitem{Beery_2018_ECCV}
Sara Beery, Grant Van~Horn, and Pietro Perona.
\newblock Recognition in terra incognita.
\newblock In {\em Proceedings of the European Conference on Computer Vision
  (ECCV)}, September 2018.

\bibitem{carlucci2019domain}
Fabio~M Carlucci, Antonio D'Innocente, Silvia Bucci, Barbara Caputo, and
  Tatiana Tommasi.
\newblock Domain generalization by solving jigsaw puzzles.
\newblock In {\em Proceedings of the IEEE Conference on Computer Vision and
  Pattern Recognition}, pages 2229--2238, 2019.

\bibitem{chabra2019stereodrnet}
Rohan Chabra, Julian Straub, Christopher Sweeney, Richard Newcombe, and Henry
  Fuchs.
\newblock Stereodrnet: Dilated residual stereonet.
\newblock In {\em Proceedings of the IEEE Conference on Computer Vision and
  Pattern Recognition}, pages 11786--11795, 2019.

\bibitem{chang2018pyramid}
Jia-Ren Chang and Yong-Sheng Chen.
\newblock Pyramid stereo matching network.
\newblock In {\em Proceedings of the IEEE Conference on Computer Vision and
  Pattern Recognition}, pages 5410--5418, 2018.

\bibitem{choi2021robustnet}
Sungha Choi, Sanghun Jung, Huiwon Yun, Joanne~T Kim, Seungryong Kim, and Jaegul
  Choo.
\newblock Robustnet: Improving domain generalization in urban-scene
  segmentation via instance selective whitening.
\newblock In {\em Proceedings of the IEEE/CVF Conference on Computer Vision and
  Pattern Recognition}, pages 11580--11590, 2021.

\bibitem{Cordts2016Cityscapes}
Marius Cordts, Mohamed Omran, Sebastian Ramos, Timo Rehfeld, Markus Enzweiler,
  Rodrigo Benenson, Uwe Franke, Stefan Roth, and Bernt Schiele.
\newblock The cityscapes dataset for semantic urban scene understanding.
\newblock In {\em Proc. of the IEEE Conference on Computer Vision and Pattern
  Recognition (CVPR)}, 2016.

\bibitem{dagaev2021too}
Nikolay Dagaev, Brett~D Roads, Xiaoliang Luo, Daniel~N Barry, Kaustubh~R Patil,
  and Bradley~C Love.
\newblock A too-good-to-be-true prior to reduce shortcut reliance.
\newblock {\em arXiv preprint arXiv:2102.06406}, 2021.

\bibitem{ganin2016domain}
Yaroslav Ganin, Evgeniya Ustinova, Hana Ajakan, Pascal Germain, Hugo
  Larochelle, Fran{\c{c}}ois Laviolette, Mario Marchand, and Victor Lempitsky.
\newblock Domain-adversarial training of neural networks.
\newblock {\em The journal of machine learning research}, 17(1):2096--2030,
  2016.

\bibitem{Geiger2012CVPR}
Andreas Geiger, Philip Lenz, and Raquel Urtasun.
\newblock Are we ready for autonomous driving? the kitti vision benchmark
  suite.
\newblock In {\em Conference on Computer Vision and Pattern Recognition
  (CVPR)}, 2012.

\bibitem{geirhos2020shortcut}
Robert Geirhos, J{\"o}rn-Henrik Jacobsen, Claudio Michaelis, Richard Zemel,
  Wieland Brendel, Matthias Bethge, and Felix~A Wichmann.
\newblock Shortcut learning in deep neural networks.
\newblock {\em arXiv preprint arXiv:2004.07780}, 2020.

\bibitem{geirhos2020}
Robert Geirhos, J{\"{o}}rn-Henrik Jacobsen, Claudio Michaelis, Richard Zemel,
  Wieland Brendel, Matthias Bethge, and Felix~A Wichmann.
\newblock {Shortcut learning in deep neural networks}.
\newblock {\em Nature Machine Intelligence}, 2(11):665--673, 2020.

\bibitem{geirhos2018imagenet}
Robert Geirhos, Patricia Rubisch, Claudio Michaelis, Matthias Bethge, Felix~A
  Wichmann, and Wieland Brendel.
\newblock Imagenet-trained cnns are biased towards texture; increasing shape
  bias improves accuracy and robustness.
\newblock {\em arXiv preprint arXiv:1811.12231}, 2018.

\bibitem{gu2020cascade}
Xiaodong Gu, Zhiwen Fan, Siyu Zhu, Zuozhuo Dai, Feitong Tan, and Ping Tan.
\newblock Cascade cost volume for high-resolution multi-view stereo and stereo
  matching.
\newblock In {\em Proceedings of the IEEE/CVF Conference on Computer Vision and
  Pattern Recognition}, pages 2495--2504, 2020.

\bibitem{guo2019group}
Xiaoyang Guo, Kai Yang, Wukui Yang, Xiaogang Wang, and Hongsheng Li.
\newblock Group-wise correlation stereo network.
\newblock In {\em Proceedings of the IEEE Conference on Computer Vision and
  Pattern Recognition}, pages 3273--3282, 2019.

\bibitem{hendrycks2019robustness}
Dan Hendrycks and Thomas Dietterich.
\newblock Benchmarking neural network robustness to common corruptions and
  perturbations.
\newblock {\em Proceedings of the International Conference on Learning
  Representations}, 2019.

\bibitem{kendall2017end}
Alex Kendall, Hayk Martirosyan, Saumitro Dasgupta, Peter Henry, Ryan Kennedy,
  Abraham Bachrach, and Adam Bry.
\newblock End-to-end learning of geometry and context for deep stereo
  regression.
\newblock In {\em Proceedings of the IEEE International Conference on Computer
  Vision}, pages 66--75, 2017.

\bibitem{lecun1998gradient}
Yann LeCun, L{\'e}on Bottou, Yoshua Bengio, and Patrick Haffner.
\newblock Gradient-based learning applied to document recognition.
\newblock {\em Proceedings of the IEEE}, 86(11):2278--2324, 1998.

\bibitem{li2018domain}
Haoliang Li, Sinno Jialin~Pan, Shiqi Wang, and Alex~C Kot.
\newblock Domain generalization with adversarial feature learning.
\newblock In {\em Proceedings of the IEEE Conference on Computer Vision and
  Pattern Recognition}, pages 5400--5409, 2018.

\bibitem{liang2018learning}
Zhengfa Liang, Yiliu Feng, Yulan Guo, Hengzhu Liu, Wei Chen, Linbo Qiao, Li
  Zhou, and Jianfeng Zhang.
\newblock Learning for disparity estimation through feature constancy.
\newblock In {\em Proceedings of the IEEE Conference on Computer Vision and
  Pattern Recognition}, pages 2811--2820, 2018.

\bibitem{liu2020stereogan}
Rui Liu, Chengxi Yang, Wenxiu Sun, Xiaogang Wang, and Hongsheng Li.
\newblock Stereogan: Bridging synthetic-to-real domain gap by joint
  optimization of domain translation and stereo matching.
\newblock In {\em Proceedings of the IEEE/CVF Conference on Computer Vision and
  Pattern Recognition}, pages 12757--12766, 2020.

\bibitem{long2015fully}
Jonathan Long, Evan Shelhamer, and Trevor Darrell.
\newblock Fully convolutional networks for semantic segmentation.
\newblock In {\em Proceedings of the IEEE conference on computer vision and
  pattern recognition}, pages 3431--3440, 2015.

\bibitem{RobotCarDatasetIJRR}
Will Maddern, Geoff Pascoe, Chris Linegar, and Paul Newman.
\newblock {1 Year, 1000km: The Oxford RobotCar Dataset}.
\newblock {\em The International Journal of Robotics Research (IJRR)},
  36(1):3--15, 2017.

\bibitem{mayer2016large}
Nikolaus Mayer, Eddy Ilg, Philip Hausser, Philipp Fischer, Daniel Cremers,
  Alexey Dosovitskiy, and Thomas Brox.
\newblock A large dataset to train convolutional networks for disparity,
  optical flow, and scene flow estimation.
\newblock In {\em Proceedings of the IEEE Conference on Computer Vision and
  Pattern Recognition}, pages 4040--4048, 2016.

\bibitem{Menze2018JPRS}
Moritz Menze, Christian Heipke, and Andreas Geiger.
\newblock Object scene flow.
\newblock {\em ISPRS Journal of Photogrammetry and Remote Sensing (JPRS)},
  2018.

\bibitem{minderer2020automatic}
Matthias Minderer, Olivier Bachem, Neil Houlsby, and Michael Tschannen.
\newblock Automatic shortcut removal for self-supervised representation
  learning.
\newblock In {\em International Conference on Machine Learning}, pages
  6927--6937. PMLR, 2020.

\bibitem{nie2019multi}
Guang-Yu Nie, Ming-Ming Cheng, Yun Liu, Zhengfa Liang, Deng-Ping Fan, Yue Liu,
  and Yongtian Wang.
\newblock Multi-level context ultra-aggregation for stereo matching.
\newblock In {\em Proceedings of the IEEE conference on computer vision and
  pattern recognition}, pages 3283--3291, 2019.

\bibitem{pan2018two}
Xingang Pan, Ping Luo, Jianping Shi, and Xiaoou Tang.
\newblock Two at once: Enhancing learning and generalization capacities via
  ibn-net.
\newblock In {\em Proceedings of the European Conference on Computer Vision
  (ECCV)}, pages 464--479, 2018.

\bibitem{pang2017cascade}
Jiahao Pang, Wenxiu Sun, Jimmy~SJ Ren, Chengxi Yang, and Qiong Yan.
\newblock Cascade residual learning: A two-stage convolutional neural network
  for stereo matching.
\newblock In {\em Proceedings of the IEEE International Conference on Computer
  Vision Workshops}, pages 887--895, 2017.

\bibitem{pensia2020extracting}
Ankit Pensia, Varun Jog, and Po-Ling Loh.
\newblock Extracting robust and accurate features via a robust information
  bottleneck.
\newblock {\em IEEE Journal on Selected Areas in Information Theory},
  1(1):131--144, 2020.

\bibitem{qiao2020learning}
Fengchun Qiao, Long Zhao, and Xi Peng.
\newblock Learning to learn single domain generalization.
\newblock In {\em Proceedings of the IEEE/CVF Conference on Computer Vision and
  Pattern Recognition}, pages 12556--12565, 2020.

\bibitem{recht2019imagenet}
Benjamin Recht, Rebecca Roelofs, Ludwig Schmidt, and Vaishaal Shankar.
\newblock Do imagenet classifiers generalize to imagenet?
\newblock In {\em International Conference on Machine Learning}, pages
  5389--5400. PMLR, 2019.

\bibitem{richter2016playing}
Stephan~R Richter, Vibhav Vineet, Stefan Roth, and Vladlen Koltun.
\newblock Playing for data: Ground truth from computer games.
\newblock In {\em European conference on computer vision}, pages 102--118.
  Springer, 2016.

\bibitem{scharstein2014high}
Daniel Scharstein, Heiko Hirschm{\"u}ller, York Kitajima, Greg Krathwohl, Nera
  Ne{\v{s}}i{\'c}, Xi Wang, and Porter Westling.
\newblock High-resolution stereo datasets with subpixel-accurate ground truth.
\newblock In {\em German conference on pattern recognition}, pages 31--42.
  Springer, 2014.

\bibitem{schoeps2017cvpr}
Thomas Sch\"ops, Johannes~L. Sch\"onberger, Silvano Galliani, Torsten Sattler,
  Konrad Schindler, Marc Pollefeys, and Andreas Geiger.
\newblock A multi-view stereo benchmark with high-resolution images and
  multi-camera videos.
\newblock In {\em Conference on Computer Vision and Pattern Recognition
  (CVPR)}, 2017.

\bibitem{Shen_2021_CVPR}
Zhelun Shen, Yuchao Dai, and Zhibo Rao.
\newblock Cfnet: Cascade and fused cost volume for robust stereo matching.
\newblock In {\em Proceedings of the IEEE/CVF Conference on Computer Vision and
  Pattern Recognition (CVPR)}, pages 13906--13915, June 2021.

\bibitem{shi2020informative}
Baifeng Shi, Dinghuai Zhang, Qi Dai, Zhanxing Zhu, Yadong Mu, and Jingdong
  Wang.
\newblock Informative dropout for robust representation learning: A shape-bias
  perspective.
\newblock In {\em International Conference on Machine Learning}, pages
  8828--8839. PMLR, 2020.

\bibitem{shi2021gradient}
Yuge Shi, Jeffrey Seely, Philip~HS Torr, N Siddharth, Awni Hannun, Nicolas
  Usunier, and Gabriel Synnaeve.
\newblock Gradient matching for domain generalization.
\newblock {\em arXiv preprint arXiv:2104.09937}, 2021.

\bibitem{tishby2015deep}
Naftali Tishby and Noga Zaslavsky.
\newblock Deep learning and the information bottleneck principle.
\newblock In {\em 2015 IEEE Information Theory Workshop (ITW)}, pages 1--5.
  IEEE, 2015.

\bibitem{tonioni2019learning}
Alessio Tonioni, Oscar Rahnama, Thomas Joy, Luigi~Di Stefano, Thalaiyasingam
  Ajanthan, and Philip~HS Torr.
\newblock Learning to adapt for stereo.
\newblock In {\em Proceedings of the IEEE Conference on Computer Vision and
  Pattern Recognition}, pages 9661--9670, 2019.

\bibitem{VolpiNSDMS18}
Riccardo Volpi, Hongseok Namkoong, Ozan Sener, John~C. Duchi, Vittorio Murino,
  and Silvio Savarese.
\newblock Generalizing to unseen domains via adversarial data augmentation.
\newblock In Samy Bengio, Hanna~M. Wallach, Hugo Larochelle, Kristen Grauman,
  Nicol{\`{o}} Cesa{-}Bianchi, and Roman Garnett, editors, {\em Advances in
  Neural Information Processing Systems 31: Annual Conference on Neural
  Information Processing Systems 2018, NeurIPS 2018, December 3-8, 2018,
  Montr{\'{e}}al, Canada}, pages 5339--5349, 2018.

\bibitem{wang2018learning}
Haohan Wang, Zexue He, Zachary~L. Lipton, and Eric~P. Xing.
\newblock Learning robust representations by projecting superficial statistics
  out.
\newblock In {\em International Conference on Learning Representations}, 2019.

\bibitem{wang2019pseudo}
Yan Wang, Wei-Lun Chao, Divyansh Garg, Bharath Hariharan, Mark Campbell, and
  Kilian~Q Weinberger.
\newblock Pseudo-lidar from visual depth estimation: Bridging the gap in 3d
  object detection for autonomous driving.
\newblock In {\em Proceedings of the IEEE Conference on Computer Vision and
  Pattern Recognition}, pages 8445--8453, 2019.

\bibitem{xu2020aanet}
Haofei Xu and Juyong Zhang.
\newblock Aanet: Adaptive aggregation network for efficient stereo matching.
\newblock {\em arXiv preprint arXiv:2004.09548}, 2020.

\bibitem{yang2019hierarchical}
Gengshan Yang, Joshua Manela, Michael Happold, and Deva Ramanan.
\newblock Hierarchical deep stereo matching on high-resolution images.
\newblock In {\em Proceedings of the IEEE Conference on Computer Vision and
  Pattern Recognition}, pages 5515--5524, 2019.

\bibitem{yang2019drivingstereo}
Guorun Yang, Xiao Song, Chaoqin Huang, Zhidong Deng, Jianping Shi, and Bolei
  Zhou.
\newblock Drivingstereo: A large-scale dataset for stereo matching in
  autonomous driving scenarios.
\newblock In {\em Proceedings of the IEEE/CVF Conference on Computer Vision and
  Pattern Recognition}, pages 899--908, 2019.

\bibitem{yang2018segstereo}
Guorun Yang, Hengshuang Zhao, Jianping Shi, Zhidong Deng, and Jiaya Jia.
\newblock Segstereo: Exploiting semantic information for disparity estimation.
\newblock In {\em Proceedings of the European Conference on Computer Vision
  (ECCV)}, pages 636--651, 2018.

\bibitem{yin2019hierarchical}
Zhichao Yin, Trevor Darrell, and Fisher Yu.
\newblock Hierarchical discrete distribution decomposition for match density
  estimation.
\newblock In {\em Proceedings of the IEEE/CVF Conference on Computer Vision and
  Pattern Recognition}, pages 6044--6053, 2019.

\bibitem{you2019pseudo}
Yurong You, Yan Wang, Wei-Lun Chao, Divyansh Garg, Geoff Pleiss, Bharath
  Hariharan, Mark Campbell, and Kilian~Q Weinberger.
\newblock Pseudo-lidar++: Accurate depth for 3d object detection in autonomous
  driving.
\newblock {\em arXiv preprint arXiv:1906.06310}, 2019.

\bibitem{yue2019domain}
Xiangyu Yue, Yang Zhang, Sicheng Zhao, Alberto Sangiovanni-Vincentelli, Kurt
  Keutzer, and Boqing Gong.
\newblock Domain randomization and pyramid consistency: Simulation-to-real
  generalization without accessing target domain data.
\newblock In {\em Proceedings of the IEEE/CVF International Conference on
  Computer Vision}, pages 2100--2110, 2019.

\bibitem{zhang2019ga}
Feihu Zhang, Victor Prisacariu, Ruigang Yang, and Philip~HS Torr.
\newblock Ga-net: Guided aggregation net for end-to-end stereo matching.
\newblock In {\em Proceedings of the IEEE Conference on Computer Vision and
  Pattern Recognition}, pages 185--194, 2019.

\bibitem{zhang2020domain}
Feihu Zhang, Xiaojuan Qi, Ruigang Yang, Victor Prisacariu, Benjamin Wah, and
  Philip Torr.
\newblock Domain-invariant stereo matching networks.
\newblock In {\em European Conference on Computer Vision}, pages 420--439.
  Springer, 2020.

\bibitem{zhang2019adaptive}
Youmin Zhang, Yimin Chen, Xiao Bai, Jun Zhou, Kun Yu, Zhiwei Li, and Kuiyuan
  Yang.
\newblock Adaptive unimodal cost volume filtering for deep stereo matching.
\newblock {\em arXiv preprint arXiv:1909.03751}, 2019.

\end{thebibliography}
}

\end{document}